# Toward Ethical Robotic Behavior
# in Human-Robot Interaction Scenarios*

Shengkang Chen
Electrical and
Computer Engineering
Georgia Tech
Atlanta GA
schen754@gatech.edu

Vidullan Surendran
Alan R. Wagner
Aerospace Engineering
Penn State University
University Park PA
vus133@psu.edu
azw78@psu.edu

Jason Borenstein
School of Public Policy
Georgia Tech
Atlanta GA
borenstein@gatech.edu

Ronald C. Arkin
Interactive Computing
Georgia Tech
Atlanta GA
arkin@gatech.edu

## ABSTRACT

This paper describes current progress on developing an ethical architecture for robots that are designed to follow human ethical decision-making processes. We surveyed both regular adults (folks) and ethics experts (experts) on what they consider to be ethical behavior in two specific scenarios: pill-sorting with an older adult and game playing with a child. A key goal of the surveys is to better understand human ethical decision-making. In the first survey, folk responses were based on the subject's ethical choices ("folk morality"); in the second survey, expert responses were based on the expert's application of different formal ethical frameworks to each scenario. We observed that most of the formal ethical frameworks we included in the survey (Utilitarianism, Kantian Ethics, Ethics of Care and Virtue Ethics) and "folk morality" were conservative toward deception in the high-risk task with an older adult when both the adult and the child had significant performance deficiencies.

## CCS CONCEPTS

• Artificial intelligence → Philosophical/theoretical foundation of artificial intelligence → Theory of mind

## KEYWORDS

Human-Robot Interaction; Robot Ethics

## 1 Introduction

Determining what counts as an "ethical" decision can be challenging, which is not only true for humans but also for robots. Yet researchers are working toward developing robots that can act ethically [1]–[3]. Various approaches have been proposed to create ethical robots including learning from moral exemplars [4] and using a set of predefined ethical rules [5]. However, these approaches may not generate appropriate behaviors in unseen and realistic environments. In particular, we want to ensure that human-robot interaction is acceptable to an end-user both in terms of experience and outcomes. To tackle these limitations, we want the actions of robots to be consistent with human ethical decision-

* Research supported by the National Science Foundation as part of the Smart and Autonomous Systems program under Grants No. 1849068 and 1848974.

making processes. Towards that end, we conducted a survey study with two separate surveys, one for folks and one for experts, to shed light on how humans make ethical decisions in two different scenarios that can involve some level of deception. This survey study is part of an ongoing NSF project [6] which aims to create an ethical architecture that can switch between different ethical frameworks to produce behaviors that are adaptable and grounded on what humans considered to be acceptable.

Deception in social robots has been an important but controversial topic. Some researchers are concerned about the undermining effect on human users [7], [8] while others believe robotic deception is permissible and can even be beneficial [9], [10]. The survey seeks to answer the question: which ethical robotic behaviors involving deception in human-robot scenarios are acceptable? Studying how people react to these situations may help answer this question.

In this study, we focused on two different scenarios: pill-sorting with an older adult and game playing with a child. Pill sorting is a common and important task for an older adult. Since incorrect sorting results can lead to serious, even fatal, consequences, we considered it a high-risk task. Moreover, the training for the task could be challenging for older adults with memory issues and can lead to frustration. We investigated whether the use of deception to keep an older adult engaged in a pill sorting task is appropriate morally, given the adult's emotions and performance history. For the game playing scenario, we chose the classic board game Connect Four. Since the outcomes of this gameplay had no obvious risk to the child, apart from frustration, we considered it a low-risk task. We chose to investigate whether an adult or a robot should let the child win intentionally to make him/her happy by either playing badly (subtle deceptions) or allowing the child to break the game's rules (cheat) in various cases.

## 2 Survey Data Collection

We collected survey data from both regular adults (folks) and ethics experts (experts). For folk survey data, we used the Amazon Mechanical Turk service and collected 100 valid responses in January 2020. For expert survey data, we invited 30 ethics experts and received 22 survey responses in February 2020. Compared with the folk survey, expert survey questions had similar wording, but the experts needed to answer the questions based on a set of formal ethical frameworks and could choose "uncertain" as a response in addition to the "yes" or "no" answer options, based on their familiarity with the framework or other factors. The "uncertain" option was not offered in the folk survey.

## 3 Synopsis of Survey Results Analysis

Given the survey data including folk responses and expert responses, we wanted to examine if there was a significant difference in the tendency to deceive between pill sorting with an older adult and game playing with a child under various ethical frameworks (including folks' opinions which we termed "folk morality"). Because the survey data are dichotomous and paired, we performed McNemar tests for large sample sizes, or exact McNemar Test for small sample sizes, using the *MLxtend* library [11]. We used these tests to determine if there are differences on whether to deceive the human subjects between the two scenarios. To study the expert survey data, we normalized the data by excluding the uncertain responses (22% in pill sorting and 38% in playing game playing). We focused on two specific cases to compare the pill sorting and the game playing scenarios:

Case 1 (minor performance deficiencies): Both the older adult and the child made one mistake in pill sorting and game playing respectively. Both were frustrated.

Case 2 (significant performance deficiencies): The older adult got half of the pill sorting task wrong, and the child lost 5 games straight. Both were also frustrated.

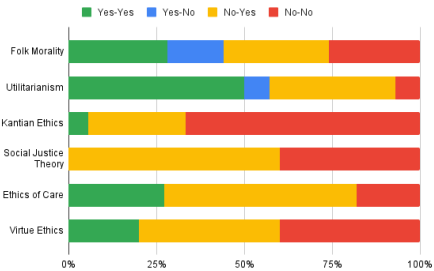

| Ethical Framework | P-value |
|---|---|
| Folk Morality | 0.0553 |
| Utilitarianism [12] | 0.2188 |
| Kantian Ethics [13] | 0.0625 |
| Social Justice Theory [14] | 0.25 |
| Ethics of Care [15] | **0.03125** |
| Virtue Ethics [16] | 0.125 |

**Figure 1. The distributions of response pairs and p-values from McNemar tests in different ethical frameworks in case 1 (both the older adult and the child made 1 mistake and were frustrated). A response pair includes a response on deception in pill sorting and a response on deception in game playing. For example, yes-no indicates the survey participant believed it was okay to deceive an older adult in pill-sorting but not for a child in game-playing.**

Based on the results of case 1 (Figure 1) we only observed significant differences (p < 0.05) between the two scenarios in the Ethics of Care framework while other ethical frameworks failed to show the statistical significances. The Ethics of Care framework indicated that it was more acceptable to deceive a child in a game than to deceive an older adult in a pill sorting task when both only made one mistake and became frustrated. However, results of case 2 (Figure 2) showed most of the ethical frameworks except Social Justice Theory had significant differences between the two scenarios when the human subjects had significant performance deficiencies and were frustrated. These results demonstrated that the ethical frameworks (folk morality, Utilitarianism, Kantian Ethics, Ethics of Care and Virtue Ethics) were significantly more restrictive regarding the use of deception in the high-risk task with an older adult than a low-risk task (game playing) with a child when the adult's or child's performance was seriously deficient. This may be because significant performance deficiencies in high-risk tasks with an older adult, in this case pill sorting, can lead to serious consequences. Comparing with results from both cases, we observed that more ethical frameworks showed significant differences between the two scenarios when those sorting pills or playing the game had much greater performance deficiencies. This observation suggests that the performance of tasks can be an important factor for each ethical framework when making decisions on whether to deceive.

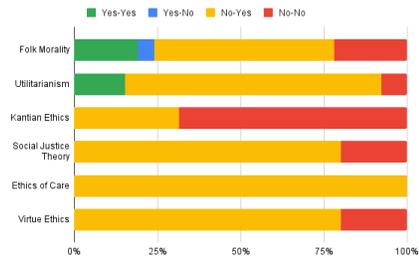

| Ethical Framework | P-value |
|---|---|
| Folk Morality | **0.0000** |
| Utilitarianism [12] | **0.002** |
| Kantian Ethics [13] | **0.0323** |
| Social Justice Theory [14] | 0.125 |
| Ethics of Care [15] | **0.001** |
| Virtue Ethics [16] | **0.0078** |

**Figure 2. The distributions of response pairs and p-values from McNemar tests in different ethical frameworks case 2 (both the older adult and the child had significant performance deficiencies and were frustrated). A response pair includes a response on deception in pill sorting and a response on deception in game playing. For example, yes-no indicates the survey participant believed it was okay to deceive an older adult in pill-sorting but not for a child in game-playing.**

## 4 Conclusion

In this abstract, we share survey data collected from both regular adults (using folk morality) and ethics experts (using formal ethical frameworks) for ethical behaviors in two specific scenarios: pill sorting with an older adult and game playing with a child. The results helped us learn more about how people perceive ethical decision making. the human subjects When the human subjects had significant performance deficiencies, ethical frameworks (folk morality, Utilitarianism, Kantian Ethics, Ethics of Care and Virtue Ethics) were significantly more likely to condemn the use of deception in the high-risk task with an older adult. There are other results which we will report regarding the risk levels and demographics, but space precludes them here.

## REFERENCES

[1] M. Anderson and S. L. Anderson, "Machine Ethics: Creating an Ethical Intelligent Agent," *AI Magazine*, vol. 28, no. 4, pp. 15–15, Dec. 2007, doi: 10.1609/AIMAG.V28I4.2065.

[2] J. S. Gordon, "Building Moral Robots: Ethical Pitfalls and Challenges," *Science and Engineering Ethics*, vol. 26, no. 1, pp. 141–157, Feb. 2020, doi: 10.1007/S11948-019-00084-5/TABLES/1.

[3] M. Scheutz and B. F. Malle, "'Think and do the right thing' - A Plea for morally competent autonomous robots," *2014 IEEE International Symposium on Ethics in Science, Technology and Engineering, ETHICS 2014*, 2014, doi: 10.1109/ETHICS.2014.6893457.

[4] D. Abel, J. MacGlashan, and M. L. Littman, "Reinforcement Learning as a Framework for Ethical Decision Making," *Workshops at the Thirtieth AAAI Conference on Artificial Intelligence*, Mar. 2016, Accessed: Jan. 25, 2022. [Online]. Available: https://www.aaai.org/ocs/index.php/WS/AAAIW16/paper/view/12582

[5] R. C. Arkin and P. Ulam, "An ethical adaptor: Behavioral modification derived from moral emotions," *Proceedings of IEEE International Symposium on Computational Intelligence in Robotics and Automation, CIRA*, pp. 381–387, 2009, doi: 10.1109/CIRA.2009.5423177.

[6] R. C. Arkin, J. Borenstein, and A. R. Wagner, "Competing ethical frameworks mediated by moral emotions in HRI: Motivations, background, and approach," Jul. 2019. doi: 10.13180/ICRES.2019.29-30.07.001.

[7] J. Danaher, "Robot Betrayal: a guide to the ethics of robotic deception," *Ethics and Information Technology 2020 22:2*, vol. 22, no. 2, pp. 117–128, Jan. 2020, doi: 10.1007/S10676-019-09520-3.

[8] A. Sharkey and N. Sharkey, "We need to talk about deception in social robotics!," *Ethics and Information Technology 2020 23:3*, vol. 23, no. 3, pp. 309–316, Nov. 2020, doi: 10.1007/S10676-020-09573-9.

[9] J. Shim and R. C. Arkin, "A taxonomy of robot deception and its benefits in HRI," *Proceedings - 2013 IEEE International Conference on Systems, Man, and Cybernetics, SMC 2013*, pp. 2328–2335, 2013, doi: 10.1109/SMC.2013.398.

[10] A. Matthias, "Robot Lies in Health Care: When Is Deception Morally Permissible?," *Kennedy Institute of Ethics journal*, vol. 25, no. 2, pp. 169–192, Jul. 2015, doi: 10.1353/KEN.2015.0007.

[11] S. Raschka, "MLxtend: Providing machine learning and data science utilities and extensions to Python's scientific computing stack," *The Journal of Open Source Software*, vol. 3, no. 24, Apr. 2018.

[12] J. Driver, "The History of Utilitarianism," in *The Stanford Encyclopedia of Philosophy*, Winter 2014., E. N. Zalta, Ed. Metaphysics Research Lab, Stanford University, 2014.

[13] R. Johnson and A. Cureton, "Kant's Moral Philosophy," in *The Stanford Encyclopedia of Philosophy*, Spring 2022., E. N. Zalta, Ed. Metaphysics Research Lab, Stanford University, 2022.

[14] J. Rawls, *A theory of justice*. Cambridge, Massachusetts : The Belknap Press of Harvard University Press, 1971.

[15] Virginia. Held, *The ethics of care personal, political, and global*. Oxford University Press, 2005.

[16] R. Hursthouse and G. Pettigrove, "Virtue Ethics," in *The Stanford Encyclopedia of Philosophy*, Winter 2018., E. N. Zalta, Ed. Metaphysics Research Lab, Stanford University, 2018.